\documentclass[11pt,a4paper]{article}
\usepackage[table]{xcolor}
\usepackage[hyperref]{emnlp-ijcnlp-2019}
\usepackage{times}
\usepackage{latexsym}
\usepackage{graphicx}
\usepackage{adjustbox}
\usepackage{url}
\usepackage{color}
\usepackage{amsmath}
\usepackage{algpseudocode}
\usepackage{multirow}
\definecolor{mycyan}{cmyk}{.3,0,0,0}
\definecolor{myred}{cmyk}{0,0.58,0.58,0}

\usepackage{pifont}
\usepackage{amsmath}
\usepackage{float}
\usepackage{multirow}
\usepackage{rotating}
\usepackage{dsfont}
\aclfinalcopy 
\usepackage{algpseudocode}

\DeclareMathOperator*{\argmax}{argmax}

\title{Entity-Consistent End-to-end Task-Oriented Dialogue System with KB Retriever}

\author{Libo Qin, Yijia Liu, Wanxiang Che\thanks{* Email corresponding.},   Haoyang Wen, Yangming Li, Ting Liu \\
	Research Center for Social Computing and Information Retrieval \\
	Harbin Institute of Technology, China \\
	{\tt \{lbqin,yjliu,car,hywen,yangmingli,tliu\}@ir.hit.edu.cn}	
}

\date{}

\begin{document}
\maketitle
\begin{abstract}
	Querying the knowledge base (KB)
	has long been a challenge in the 
	end-to-end task-oriented dialogue system.
	Previous sequence-to-sequence (Seq2Seq) dialogue generation work
	treats the KB query as an attention over the entire KB,
	without the guarantee that the generated entities
	are consistent with each other.
	In this paper, we propose a novel framework which queries the KB in two steps to improve the consistency of generated entities.
	In the first step, 	inspired by the observation
	that a response can usually be supported by a single KB row,
	we introduce a KB retrieval component
	which explicitly returns the most relevant KB row given a dialogue history.
	The retrieval result is further used to filter the irrelevant entities
	in a Seq2Seq response generation model
	to improve the consistency among the output entities.
	In the second step, we further perform the attention mechanism to address the most correlated KB column.
	Two methods are proposed to make the training feasible 
	without labeled retrieval data, which include distant supervision and Gumbel-Softmax technique.
	Experiments on 
	two publicly available task oriented dialog datasets show
	the effectiveness of our model
	by outperforming the baseline systems 
	and producing entity-consistent responses.
\end{abstract}

\section{Introduction}
	\textit{Task-oriented dialogue system}, which helps users to achieve specific goals with natural language,
is attracting more and more research attention.
With the success of the sequence-to-sequence (Seq2Seq) models 
in text generation \cite{sutskever2014sequence,bahdanau2014neural,luong-pham-manning:2015:EMNLP,K16-1028,nallapati2016sequence},
several works tried to model the task-oriented dialogue as the Seq2Seq generation of response
from the dialogue history \cite{eric-manning:2017:EACL,eric:2017:SIGDial,madotto2018mem2seq}.
This kind of modeling scheme frees the task-oriented dialogue system from
the manually designed pipeline modules 
and heavy annotation labor for these modules.
	\begin{figure}[t]
	\centering
	\begin{adjustbox}{width=0.48\textwidth}
		\begin{tabular}{|l|l|l|l|l|}
			\hline\textbf{Address} & \textbf{Distance} & \textbf{POI type} & \textbf{POI} & \textbf{Traffic info} \\\hline
			638 Amherst St&3 miles&grocery store&Sigona Farmers Market&car collision nearby\\
			269 Alger Dr&1 miles&coffee or tea place&Cafe Venetia&car collision nearby\\
			5672 barringer street & 5 miles & certain address & 5672 barringer street & no traffic \\
			\rowcolor{mycyan}200 Alester Ave&2 miles&gas station&Valero&road block nearby\\
			\cellcolor{myred}899 Ames Ct&5 miles&hospital&Stanford Childrens Health&moderate traffic\\
			481 Amaranta Ave&1 miles&parking garage&Palo Alto Garage R&moderate traffic\\
			145 Amherst St&1 miles&coffee or tea place&Teavana&road block nearby\\
			409 Bollard St&5 miles&grocery store&Willows Market&no traffic\\\hline
		\end{tabular}
	\end{adjustbox}
	\newline
	\vspace{0.25em}
	\newline
	\begin{adjustbox}{width=0.48\textwidth}
		\begin{tabular}{ll}
			\textbf{Driver} & Address to the gas station.\\
			\textbf{Car} & Valero is located at 200 Alester Ave.\\
			\textbf{Driver} & OK , please give me directions via a route that avoids all heavy traffic.\\
			\textbf{Car} & Since there is a road block nearby, I found another route for you and I sent it on your screen.\\
			\textbf{Driver} & Awesome thank you.\\ 
		\end{tabular}
	\end{adjustbox}
	\caption{
		An example of a task-oriented dialogue that incorporates a knowledge base (KB). 
		The fourth row in KB supports the second turn of the dialogue. 
		A dialogue system will produce a response with \textit{conflict entities} if it includes the \textcolor{cyan}{POI} in the fourth row
		and the \textcolor{red}{address} in the fifth row, like ``\textit{Valero is located at 899 Ames Ct}''.
	}
	\label{fig:example}
\end{figure}
Different from typical text generation, 
the successful conversations for task-oriented dialogue system heavily depend on accurate
knowledge base (KB) queries.
Taking the dialogue in Figure \ref{fig:example} as an example, 
to answer the driver's query on the gas station, the dialogue system is required to
retrieve the entities 
like ``\texttt{200 Alester Ave}'' and ``\texttt{Valero}''.
For the task-oriented system based on Seq2Seq generation, 
there is a trend in recent study towards 
modeling the KB
query as an attention network over
the entire KB entity representations,
hoping to learn a model to pay more attention to the relevant entities \cite{eric:2017:SIGDial,madotto2018mem2seq,reddy2018multi,wen2018sequence}.

Though achieving good end-to-end dialogue generation with over-the-entire-KB attention mechanism,
these methods do not guarantee
the generation consistency regarding KB entities
and sometimes yield responses with conflict entities,
like ``\textit{Valero is located at 899 Ames Ct}'' for the gas station query (as shown in Figure \ref{fig:example}). 
In fact, the correct address for \textit{Valero} is \textit{200 Alester Ave}.
A consistent response is relatively easy
to achieve for the conventional pipeline systems
because they query the KB by issuing API calls \cite{bordes-weston:2017:ICLR,wen:2017:EACL,wen:2017:ICML},
and the returned 
entities, which typically come from a single KB row,
are consistently related
to the object (like the ``gas station'')
that serves the user's request.
This indicates that \textit{a response can usually be supported by a single KB row}.
It's promising to incorporate such observation into
the Seq2Seq dialogue generation model,
since it encourages KB relevant generation and 
avoids the model from
producing responses with conflict entities.

To achieve entity-consistent generation in the Seq2Seq task-oriented dialogue system,
we propose a novel framework which query the KB in two steps.
In the first step, we introduce a retrieval module --- KB-retriever to explicitly query the KB.
Inspired by the observation that a single KB row usually supports a response,
given the dialogue history and a set of KB rows,
the KB-retriever uses a memory network \cite{sukhbaatar2015end}
to select the
most relevant row.
The retrieval result is then fed into a Seq2Seq dialogue generation model 
to filter the irrelevant KB entities and improve the consistency within the generated entities.
In the second step, we further perform attention mechanism to address the most correlated KB column.
Finally, we adopt the copy mechanism to incorporate the retrieved KB entity.

Since dialogue dataset is not typically annotated with the retrieval results, 
training the KB-retriever is non-trivial.
To make the training feasible, we
propose two methods:
1) we use a set of heuristics to derive the training data
and train the retriever in a distant supervised fashion;
2) we use Gumbel-Softmax \cite{45822} 
as an approximation of the non-differentiable selecting process
and train the retriever along with 
the Seq2Seq dialogue generation model. 
Experiments on two publicly available datasets ({\bf{Camrest}} \cite{wen:2017:EACL} and {\bf{InCar Assistant}} \cite{eric:2017:SIGDial}) 
confirm
the effectiveness of the KB-retriever.
Both the retrievers
trained with distant-supervision and Gumbel-Softmax technique outperform
the compared systems
in the automatic and human evaluations. 
Analysis empirically verifies our assumption
that more than 80\% responses
in the dataset can be supported by a single KB row
and better retrieval results lead to better task-oriented dialogue generation performance.

\section{Definition}
In this section,
we will describe the input and output of the end-to-end task-oriented dialogue system,
and the definition of Seq2Seq task-oriented dialogue generation.

\subsection{Dialogue History}
Given a dialogue between a user ($u$) and a system ($s$),
we follow \newcite{eric:2017:SIGDial} and
represent the $k$-turned \textit{dialogue utterances} as 
$\{(u_{1}, s_{1} ), (u_{2} , s_{2} ), ... , (u_{k}, s_{k})\}$.
At the $i^{\text{th}}$ turn of the dialogue, we aggregate dialogue context
which consists of the tokens of $(u_{1}, s_{1}, ..., s_{i-1}, u_{i})$ and use $\mathbf{x} = (x_{1}, x_{2}, ..., x_{m})$ 
to denote the whole \textit{dialogue history} word by word,
where $m$ is the number of tokens in the dialogue history.
\begin{figure*}[t]
	\centering
	\includegraphics[scale=0.32]{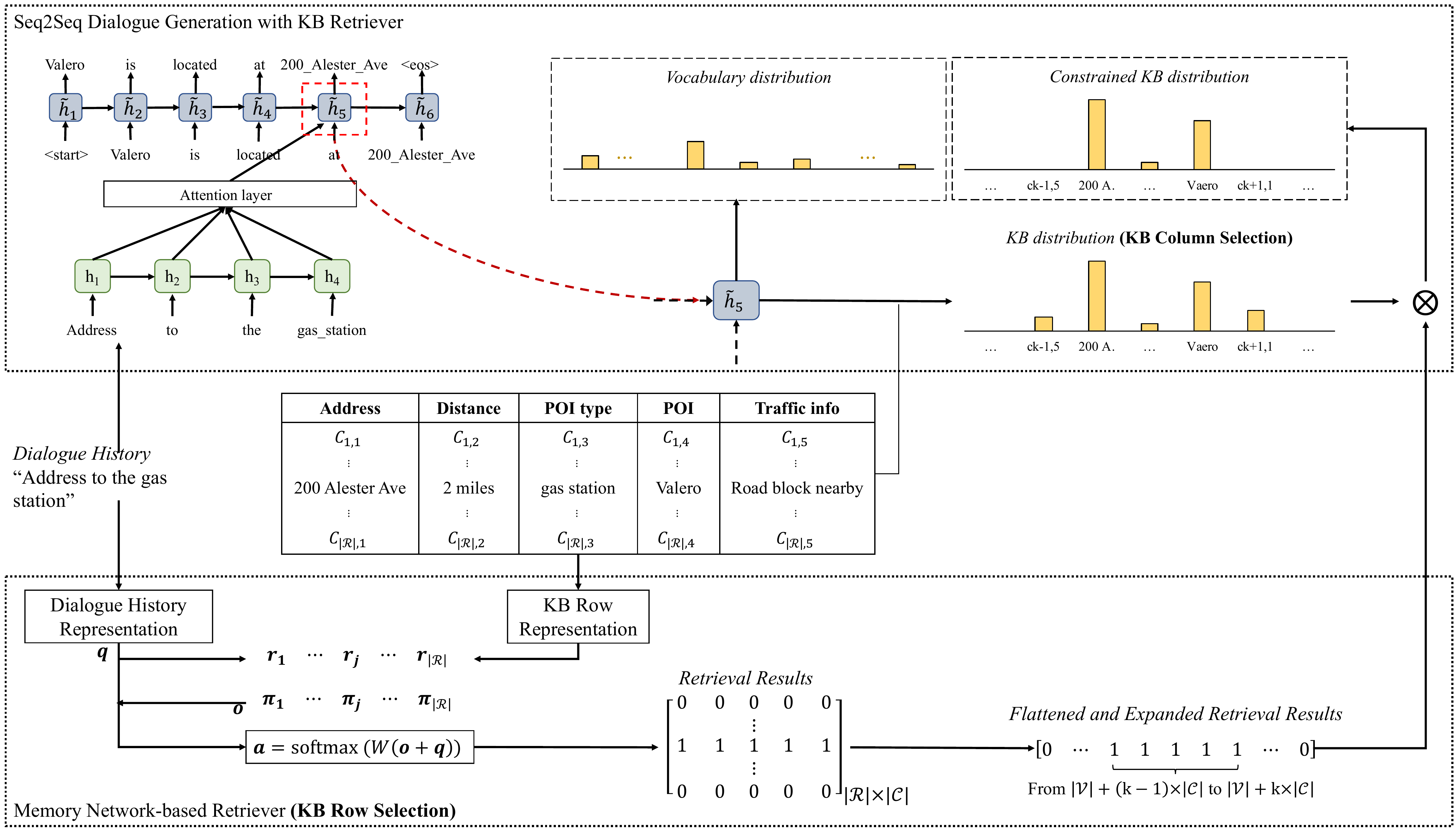}
	\caption{
		The workflow of our Seq2Seq task-oriented dialogue generation model with KB-retriever.
		For simplification, we draw the single-hop memory network instead of the multiple-hop one we use in our model.
	}
	\label{fig:framework}
\end{figure*}

\subsection{Knowledge Base}
In this paper, we assume to 
have the 
access 
to a relational-database-like 
KB  $B$,
which consists of $|\mathcal{R}|$ rows and $|\mathcal{C}|$ columns.
The value of entity in the $j^{\text{th}}$ row and the $i^{\text{th}}$ column is noted as $v_{j, i}$.

\subsection{Seq2Seq Dialogue Generation}
We define the Seq2Seq task-oriented dialogue generation
as finding the most likely response $\mathbf{y}$
according to the input dialogue history $\mathbf{x}$ and KB $B$.
Formally, the probability of a response is defined as
\[
p(\mathbf{y} \mid \mathbf{x}, B) = \prod_{t=1}^{n}  p (y_t \mid  y_1, ..., y_{t - 1}, \mathbf{x}, B),
\]
where $y_t$ represents an output token.

\section{Our Framework}
In this section, we describe our framework for end-to-end task-oriented dialogues.
The architecture of our framework is demonstrated in Figure ~\ref{fig:framework},  
which consists of two major components including an memory network-based retriever and the seq2seq dialogue generation with KB Retriever.
Our framework first uses the  KB-retriever to select the most relevant KB row and further filter the irrelevant entities
in a Seq2Seq response generation model
to improve the consistency among the output entities.
While in decoding, we further perform the attention mechanism to choose the most probable KB column.
We will present the details of our framework in the following sections.

\subsection{Encoder}
In our encoder, we adopt the bidirectional LSTM 
\cite[BiLSTM]{hochreiter1997long} to encode the dialogue history  $\mathbf{x}$, 
which captures temporal relationships within the sequence.
The encoder first map the tokens in $\mathbf{x}$ to vectors
with embedding function $\phi^{\text{emb}}$,
and then the BiLSTM read the vector forwardly and backwardly to produce context-sensitive  hidden states $(\mathbf{h}_{1}, \mathbf{h}_2, ..., \mathbf{h}_{m})$
by repeatedly applying the recurrence 
$\mathbf{h}_{i}=\text{BiLSTM}\left( \phi ^{\text{emb}}\left( x_{i}\right) , \mathbf{h}_{i-1}\right)$.

\subsection{Vanilla Attention-based Decoder}\label{sec:method:vanilla-decoder}

Here, we follow \newcite{eric:2017:SIGDial} to adopt the attention-based decoder to generation the response word by word.
LSTM is also used to represent the partially generated output sequence $(y_{1}, y_2, ...,y_{t-1})$
as $(\tilde{\mathbf{h}}_{1}, \tilde{\mathbf{h}}_2, ...,\tilde {\mathbf{h}}_t)$.
For the generation of next token $y_t$, their model
first calculates an attentive representation $\tilde{\mathbf{h}}^{'}_t$ of the dialogue history as
\begin{align*}
\mathbf{u}^{t}_{i} & = W_{2}\  \tanh(W_{1}\ [\mathbf{h}_{i}, \tilde {\mathbf{h}_{t}}]), \\
\mathbf{a}^t_{i} & = \text{softmax} (\mathbf{u}^t_{i}), \\
\tilde{\mathbf{h}}^{'}_t & = \sum ^{m}_{i=1} \mathbf{a}^t_{i}\cdot \mathbf{h}_{i}.
\end{align*}
Then, the concatenation of the hidden representation of the partially outputted sequence $\tilde{\mathbf{h}}_t$
and the attentive dialogue history representation $\tilde{\mathbf{h}}^{'}_t$
are projected to the vocabulary space $\mathcal{V}$ by $U$ as
\[\mathbf{o}_t = U\ [\tilde{\mathbf{h}}_t, \tilde{\mathbf{h}}^{'}_t], \]
to calculate the score (logit) for the next token generation.
The probability of next token $y_t$  is finally calculated as
\[p(y_t \mid y_1, ..., y_{t-1}, \mathbf{x}, B) = \text{softmax}(\mathbf{o}_t).\]

\subsection{Entity-Consistency Augmented Decoder}
As shown in section \ref{sec:method:vanilla-decoder}, we can see that the generation of tokens are just based on the dialogue history attention, which makes the model ignorant to the KB entities.
In this section, we present how to query the KB explicitly in two steps for improving the entity consistence,
which first adopt the KB-retriever to select the most relevant KB row and the generation of KB entities from the entities-augmented decoder is constrained to the entities within the most probable row, thus improve the entity generation consistency.
Next, we perform the column attention to select the most probable KB column.
Finally, we show how to use the copy mechanism to incorporate the retrieved entity while decoding.
\subsubsection{KB Row Selection}\label{model:row_retriever}
In our framework, our KB-retriever takes the dialogue history and KB rows as inputs and selects the most
 relevant row. This selection process resembles the task of selecting one word from 
 the inputs to answer questions \cite{sukhbaatar2015end}, and we use a memory network to model this process.
In the following sections, we will first describe how to represent the inputs, then we will talk about our memory network-based retriever

\paragraph{Dialogue History Representation:}
We encode the dialogue history by adopting the neural bag-of-words (BoW) followed the original paper \cite{sukhbaatar2015end}.
Each token in the dialogue history is mapped into a vector by another embedding function $\phi^{\text{emb}'}(x)$ 
and the dialogue history representation $\mathbf{q}$ is computed as the sum of these vectors:
$\mathbf{q} = \sum ^{m}_{i=1} \phi^{\text{emb}'} (x_{i}) $.

\paragraph{KB Row Representation:}
In this section, we describe how to encode the KB row. 
Each KB cell is represented as the 
cell value $v$ embedding as $\mathbf{c}_{j, k} = \phi^{\text{value}}(v_{j, k})$,
and the neural BoW is also used 
to represent a KB row $\mathbf{r}_{j}$ as
$\mathbf{r}_{j} = \sum_{k=1}^{|\mathcal{C}|} \mathbf{c}_{j,k}$.

\paragraph{Memory Network-Based Retriever:}
We model the KB retrieval process as selecting the row that
most-likely supports the response generation.
Memory network \cite{sukhbaatar2015end} has shown to be effective to
model this kind of selection.
For a $n$-hop memory network,
the model keeps a set of input matrices $\{R^{1}, R^{2}, ..., R^{n+1}\}$,
where each $R^{i}$ is a stack of $|\mathcal{R}|$ inputs $(\mathbf{r}^{i}_1, \mathbf{r}^{i}_2, ..., \mathbf{r}^{i}_{|\mathcal{R}|})$.
The model also keeps query $\mathbf{q}^{1}$ as the input.
A single hop memory network
computes the probability $\mathbf{a}_j$ of selecting 
the $j^{\text{th}}$ input as
\begin{align*}
\boldsymbol{\pi}^{1} & = \text{softmax}((\mathbf{q}^{1})^{T}\ R^{1}), \\
\mathbf{o}^{1} & = \sum_i \boldsymbol{\pi}^{1}_i \mathbf{r}^{2}_i, \\
\mathbf{a} & =  \text{softmax}{(W^{\text{mem}} \ (\mathbf{o}^{1} + \mathbf{q}^{1}))}.
\end{align*}
For the multi-hop cases, layers of single hop memory network
are stacked and the query of the $(i+1)^{\text{th}}$ layer network
is computed as 
\[\mathbf{q}^{i+1} = \mathbf{q}^{i} + \mathbf{o}^{i},\]
and the output of the last layer is used
as the output of the whole network.
For more details about memory network, please refer to the original paper \cite{sukhbaatar2015end}.

After getting $\mathbf{a}$, we represent the retrieval results
as a 0-1 matrix $T \in \{0, 1\}^{|\mathcal{R}|\times \mathcal{|C|}}$, where each element in $T$
is calculated as
\begin{equation}\label{eq:decode}
T_{j, *} = \mathds{1}[j = \argmax_i \mathbf{a}_i].
\end{equation}
In the retrieval result, $T_{j, k}$ indicates whether the entity in the $j^{\text{th}}$ row and the $k^{\text{th}}$
column is relevant to the final generation of the response.
In this paper, we further flatten T to a 0-1 vector $\mathbf{t} \in \{0, 1\}^{|\mathcal{E}|}$ (where $|\mathcal{E}|$ equals $|\mathcal{R}|\times \mathcal{|C|}$) as our retrieval row results.

\subsubsection{KB Column Selection}
After getting the retrieved row result that indicates which KB row is the most relevant to the generation, 
we further perform column attention in decoding time to select the probable KB column.
For our KB column selection, following the \newcite{eric:2017:SIGDial} we use the decoder hidden state $(\tilde{\mathbf{h}}_{1}, \tilde{\mathbf{h}}_2, ...,\tilde {\mathbf{h}}_t)$ to compute an attention score with the embedding of column attribute name.
The attention score  $\mathbf{c}\in R^{|\mathcal{E}|}$ then become the logits of the column be selected, which can be calculated as
\begin{align*}
\mathbf{c}_{j} & = W^{'}_{2}\  \tanh(W^{'}_{1}\ [\mathbf{k}_{j}, \tilde {\mathbf{h}_{t}}]),
\end{align*}
where $\mathbf{c}_j$ is the attention score of the $j^{\text{th}}$ KB column, $\mathbf{k}_j$ is represented with the embedding of word embedding of KB column name. $W^{'}_{1}$, $W^{'}_{2}$ and $\mathbf{t}^{T}$ are trainable parameters of the model.
 
 \subsubsection{Decoder with Retrieved Entity }
 After the row selection and column selection, 
 we can define the final retrieved KB entity score as the element-wise dot between the row retriever result and the column selection score, which can be calculated as

\begin{equation} \label{eq:retriever}
	  \mathbf{v}^{t}  = \mathbf{t} * \mathbf{c},
\end{equation}
 where the $v^{t}$  indicates the final KB retrieved entity score.
 Finally, we follow	\newcite{eric:2017:SIGDial} to use copy mechanism to incorporate the retrieved entity, which can be defined as 
 \[\mathbf{o}_t = U\ [\tilde{\mathbf{h}}_t, \tilde{\mathbf{h}^{'}}_t] + \mathbf{v}^t,\] 
 where $\mathbf{o}_t$’s dimensionality is $ |\mathcal{V}|$ +$|\mathcal{E}|$. In $\mathbf{v}^t$ , lower $ |\mathcal{V}|$ is zero and the rest$|\mathcal{E}|$ is retrieved entity scores.

 \section{Training the KB-Retriever}
 As mentioned in section~ \ref{model:row_retriever}, we adopt the memory network to train our KB-retriever.
 However, in the Seq2Seq dialogue generation,
 the training data does not include the
 annotated KB row retrieval results, which makes
 supervised
 training the KB-retriever impossible.
 To tackle this problem,
 we propose two training methods for our KB-row-retriever.
 1) In the first method, inspired by the recent success of
 distant supervision in information extraction \cite{zeng2015distant,mintz2009distant,N13-1095,xu2013filling},
 we take advantage of the similarity between the surface string
 of KB entries and the reference response, and
 design a set of heuristics to extract training data for the KB-retriever.
 2) In the second method, instead of training the KB-retriever as
 an independent component, we train it along with the
 training of the Seq2Seq dialogue generation.
 To make the retrieval process in Equation \ref{eq:decode} differentiable,
 we use Gumbel-Softmax \cite{45822} as an approximation of the $\argmax$ during training.
 
 \subsection{Training with Distant Supervision}
 Although it's difficult to obtain
 the annotated retrieval data for the KB-retriever,
 we can ``guess'' the most relevant KB row from
 the reference response,
 and then obtain the weakly labeled data for the retriever.
Intuitively, for the current utterance in the same dialogue which usually belongs to one topic and the KB row that contains the largest number of
entities mentioned in the whole dialogue should support the utterance.
 In our training with distant supervision, 
 we further simplify our assumption and 
 assume that
 one dialogue which is usually belongs to one topic and can be supported by the most relevant KB row,
 which means for a $k$-turned dialogue, we construct
 $k$ pairs of training instances for the retriever and all
 the inputs $(u_{1}, s_{1}, ..., s_{i-1}, u_{i} \mid i \le k)$ are associated
 with the same weakly labeled KB retrieval result $T^*$.
 
 In this paper, we compute each row's similarity to the whole dialogue
 and choose the most similar row as $T^*$.
 We define the similarity of each row
 as the number of matched spans
 with the surface
 form of the entities in the row.
 Taking the dialogue in Figure \ref{fig:example} for an example,
 the similarity of the 4$^\text{th}$ row equals to 4 with ``\texttt{200 Alester Ave}'',
 ``\texttt{gas station}'', ``\texttt{Valero}'', and ``\texttt{road block nearby}''
 matching the dialogue context;
 and the similarity of the 7$^\text{th}$ row equals to 1 with only ``\texttt{road block nearby}''
 matching.
 
 In our model with the distantly supervised retriever,
 the retrieval results serve as the input
 for the Seq2Seq generation.
 During training the Seq2Seq generation,
 we use the weakly labeled retrieval result $T^{*}$
 as the input.

 \subsection{Training  with Gumbel-Softmax}
 In addition to treating the row retrieval result as an input to the generation model,
 and training the kb-row-retriever independently,
 we can train it along with the
 training of the Seq2Seq dialogue generation in an end-to-end fashion.
 The major difficulty of such a training scheme
 is that the discrete retrieval result is not differentiable
 and the training signal from the generation model cannot be passed
 to the parameters of the retriever.
 Gumbel-softmax technique \cite{45822} has been shown an
 effective approximation to the discrete variable
 and proved to work in sentence representation.
 In this paper, we adopt the Gumbel-Softmax technique to train the KB retriever.
 We use \[
 T^{\text{approx}}_{j,*} = \frac{\text{exp}{ ((\log(\mathbf{a}_j) + \mathbf{g}_{j})/\tau) } }{\sum_i{\text{exp} {((\log(\mathbf{a}_i) + \mathbf{g}_{i})/\tau})}},
 \]
 as the approximation of $T$,
 where
 $\mathbf{g}_{j}$ are i.i.d samples drawn from $\text{Gumbel}(0,1)$\footnote{
 	We sample $\mathbf{g}$ by drawing $u \sim \text{Uniform}(0, 1)$ then computing $\mathbf{g} = -\log(-\log(u))$.
 }
 and $\tau$ is a constant that controls the smoothness of the distribution.
 $T^{\text{approx}}_{j}$ replaces $T^{\text{}}_{j}$ in equation~\ref{eq:decode} and goes through the same
 flattening and expanding process as $\mathbf{V}$
 to get $\mathbf{v}^{\mathbf{t}^{\text{approx}'}}$ and
 the training signal from Seq2Seq generation
 is passed via the logit
 \[
 \mathbf{o}^{\text{approx}}_t = U\ [\tilde{\mathbf{h}}_t, \tilde{\mathbf{h}}^{'}_t] +\mathbf{v}^{\mathbf{t}^{\text{approx}'}}.
 \]	
 To make training with Gumbel-Softmax more stable,
 we first initialize the parameters by pre-training the KB-retriever with distant supervision and further fine-tuning our framework.
 
 \subsection{Experimental Settings}
We choose the InCar Assistant dataset \cite{eric:2017:SIGDial} including three distinct domains: navigation, weather and calendar domain. 
For weather domain, we follow \newcite{wen2018sequence} to separate the highest temperature, lowest temperature and weather attribute into three different columns.
For calendar domain, there are some dialogues without a KB or incomplete KB.
In this case, we padding a special token ``-'' in these incomplete KBs.
Our framework is trained separately in these three domains, using the same train/validation/test split sets as \newcite{eric:2017:SIGDial}.\footnote{We obtain the BLEU and Entity F1 score on the whole InCar dataset by mixing all generated response and evaluating them together.} 
To justify the generalization of the proposed model, we also use another public CamRest dataset \cite{wen:2017:EACL} and partition the datasets into training, validation and testing set in the ratio 3:1:1.\footnote{The dataset can be available at: \url{https://github.com/yizhen20133868/Retriever-Dialogue}}
Especially, we hired some human experts to format the CamRest dataset by equipping the corresponding KB to every dialogues. 
 
 All hyper-parameters are selected according
 to validation set. We use a three-hop memory network to model
 our KB-retriever.
 The dimensionalities of the embedding is selected from $\{100, 200\}$ and LSTM hidden units is selected from $\{50, 100, 150, 200, 350\}$.
 The dropout we use in our framework is selected from $\{0.25, 0.5, 0.75\}$ and the batch size we adopt is selected from  $\{1,2\}$.
 L2 regularization is used on our model with a tension of $5\times 10^{-6}$ for reducing overfitting.
 For training the retriever with distant supervision, we adopt the weight typing trick \cite{E17-1001}.
 We use Adam \cite{kingma-ba:2014:ICLR} to optimize the parameters in our model and 
 adopt the suggested hyper-parameters for optimization.
 
 We adopt both the automatic and human evaluations in our experiments.

 \subsection{Baseline Models}
 We compare our model with several baselines including:
 \begin{itemize}
 	\item \textbf{Attn seq2seq}  \cite{luong-etal-2015-effective}:
 	 A model with simple attention over the input context at each time step during decoding.
 	 	\item \textbf{Ptr-UNK}  \cite{P16-1014}:
 	 Ptr-UNK is the model which augments a sequence-to-sequence architecture with attention-based copy mechanism over the encoder context.

 	\item \textbf{KV Net}  \cite{eric:2017:SIGDial}:
     The model adopted and
 	argumented decoder which decodes
 	over the concatenation of vocabulary and KB entities,
 	which allows the model to generate entities.
 	\item \textbf{Mem2Seq} \cite{madotto2018mem2seq}:
 	Mem2Seq is the model that takes dialogue history and KB entities
 	as input and uses a pointer gate
 	to control either generating a vocabulary word or selecting 
 	an input as the output.
 	\item \textbf{DSR} \cite{wen2018sequence}:  DSR leveraged dialogue state representation to retrieve the KB implicitly and  applied copying mechanism to retrieve entities from knowledge base while decoding.

 \end{itemize}
 In InCar dataset, 
  for the \textit{Attn seq2seq},  \textit{Ptr-UNK} and \textit{Mem2seq}, we adopt the reported results from \newcite{madotto2018mem2seq}.
  In CamRest dataset, for the \textit{Mem2Seq}, we adopt their open-sourced code
  to get the results while for the \textit{DSR}, we run their code on the same dataset to obtain the results.\footnote{We adopt the same pre-processed dataset from \newcite{madotto2018mem2seq}. We can find that experimental results is slightly different with their reported performance \cite{wen2018sequence} because of their different tokenized utterances and normalization for entities.}
  
\begin{table*}[ht]
	\centering
			\begin{adjustbox}{width=0.85\textwidth}
	\begin{tabular}{l||ccccc||ll}
		\hline
		& \multicolumn{5}{c}{InCar}                                                                                                                 & \multicolumn{2}{c}{CamRest} \\
		\hline \hline
		Model                                                           & BLEU & F1 & \begin{tabular}[c]{@{}l@{}}Navigate\\~ ~~ F1\end{tabular} & \begin{tabular}[c]{@{}c@{}}Weather \\F1\end{tabular} & \begin{tabular}[c]{@{}c@{}}Calendar \\F1\end{tabular} & BLEU & F1                             \\ 
		\hline
		
		{Human}*  \cite{eric:2017:SIGDial}                                                     &      13.5& 60.7   &     55.2                                                      &                     61.6                                 &             64.3&  -    &   -                            \\
	{Rule-Based}*  \cite{eric:2017:SIGDial}                                                     &      6.6& 43.8   &     40.4                                                      &                     39.5                                 &             61.3&  -    &   -                            \\
		KV Net*  \cite{eric:2017:SIGDial}                                                     &      13.2& 48.0   &     41.3                                                      &                     47.0                                 &             62.9&  -    &   -                            \\
		\hline
		
		Attn seq2seq \cite{luong-etal-2015-effective}                                                  & 9.3      & 11.9  & 10.8                                                &     25.6                                                 &   23.4          & -    & -                                 \\
		Ptr-UNK \cite{P16-1014}                                                  & 8.3      & 22.7  & 14.9                                                &     26.7                                                 &   26.9          & -    & -                                  \\
		Mem2Seq    \cite{madotto2018mem2seq}                                                        & 12.6    & 33.4   &                20.0                                           &        32.8                                              & 49.3             & 16.6    &   42.4                             \\
		DSR
		\cite{wen2018sequence}                                             &  12.7    & 51.9   &      52.0                                                     & 50.4                                                     & 52.1            &   18.3   &    53.6                     \\ 
		\hline
		w/ distant supervision  & {\bf 14.1}   & {51.9}     & {51.6}   & {49.6}                                                         &    {54.2}                                                 &  17.4  &58.0                                 \\  
		w/ Gumbel-Softmax & {13.9}   & {\bf 53.7}     & {\bf 54.5}   & {\bf 52.2}                                                         &    {\bf 55.6}                                                 & {\bf 18.5}    &\textbf{58.6 }                                  \\   \hline                  
	\end{tabular}
\end{adjustbox}
	\caption{Comparison of our model with baselines} \label{tab:results}
	\vspace{-0.3cm}
\end{table*}

\section{Results}
  Follow the prior works \cite{eric:2017:SIGDial,madotto2018mem2seq,wen2018sequence}, we adopt the \textit{BLEU} and the \textit{Micro Entity F1} to evaluate our model performance.
The experimental results are illustrated in Table~ \ref{tab:results}. 
 
 In the first block of Table~\ref{tab:results}, we show the Human, rule-based and KV Net (with*) result which are reported from \newcite{eric:2017:SIGDial}.
We argue that their results are not directly comparable because their work uses the entities in thier canonicalized forms, which are not calculated based on real entity value. 
It's noticing that our framework with two methods still outperform KV Net in InCar dataset on whole BLEU and Entity F metrics, which demonstrates the effectiveness of our framework.
 
 In the second block of Table~\ref{tab:results}, we can see that our framework trained with both the distant supervision and the Gumbel-Softmax beats all existing models on two datasets.
 Our model outperforms each baseline on both BLEU and F1 metrics.
 In InCar dataset, Our model with Gumbel-Softmax has the highest BLEU
 compared with baselines, which 
 which shows that our framework can generate more fluent response.
 Especially, our framework has achieved 2.5\% improvement on navigate domain, 1.8\% improvement on weather domain and 3.5\% improvement on calendar domain on F1 metric. It indicates that the 
 effectiveness of our KB-retriever module and our framework can 
 retrieve more correct entity from KB.
 In CamRest dataset, the same trend of improvement has been witnessed, which further show the effectiveness of our framework.
 
 Besides, we observe that the model trained with Gumbel-Softmax outperforms with distant supervision method. 
 We attribute this to the fact that the KB-retriever and the Seq2Seq module are fine-tuned in an end-to-end fashion, which can refine the KB-retriever and further promote the dialogue generation.
 	\subsection{{The proportion of responses that can be supported by a single KB row}}
 	In this section, we verify
 	our assumption by examining the
 	proportion of responses that can be supported by a single row.
 
 We define a response being supported by the most relevant KB row
 as all the responded entities are included by
 that row.
 We study the proportion of these responses
 over the test set.
 The number is 95\%
 for the navigation domain, 90\% for the CamRest dataset
 and 80\% for the weather domain.
 This confirms our assumption
 that most responses can be supported by the relevant KB row.
 Correctly retrieving the supporting row
 should be beneficial.
 
 We further study the weather domain
 to see the rest 20\% exceptions.
 Instead of being supported by multiple rows,
 most of these exceptions cannot be supported by any KB row.
 For example, there is one case whose reference response is ``\textit{It 's not rainy today}'',
 and the related KB entity is \texttt{sunny}.
 These cases provide challenges beyond the scope of this paper.
 If we consider this kind of cases as being supported by a single row,
 such proportion in the weather domain is 99\%.

\begin{figure}[t!]
	\centering
	\includegraphics[width=0.8\columnwidth]{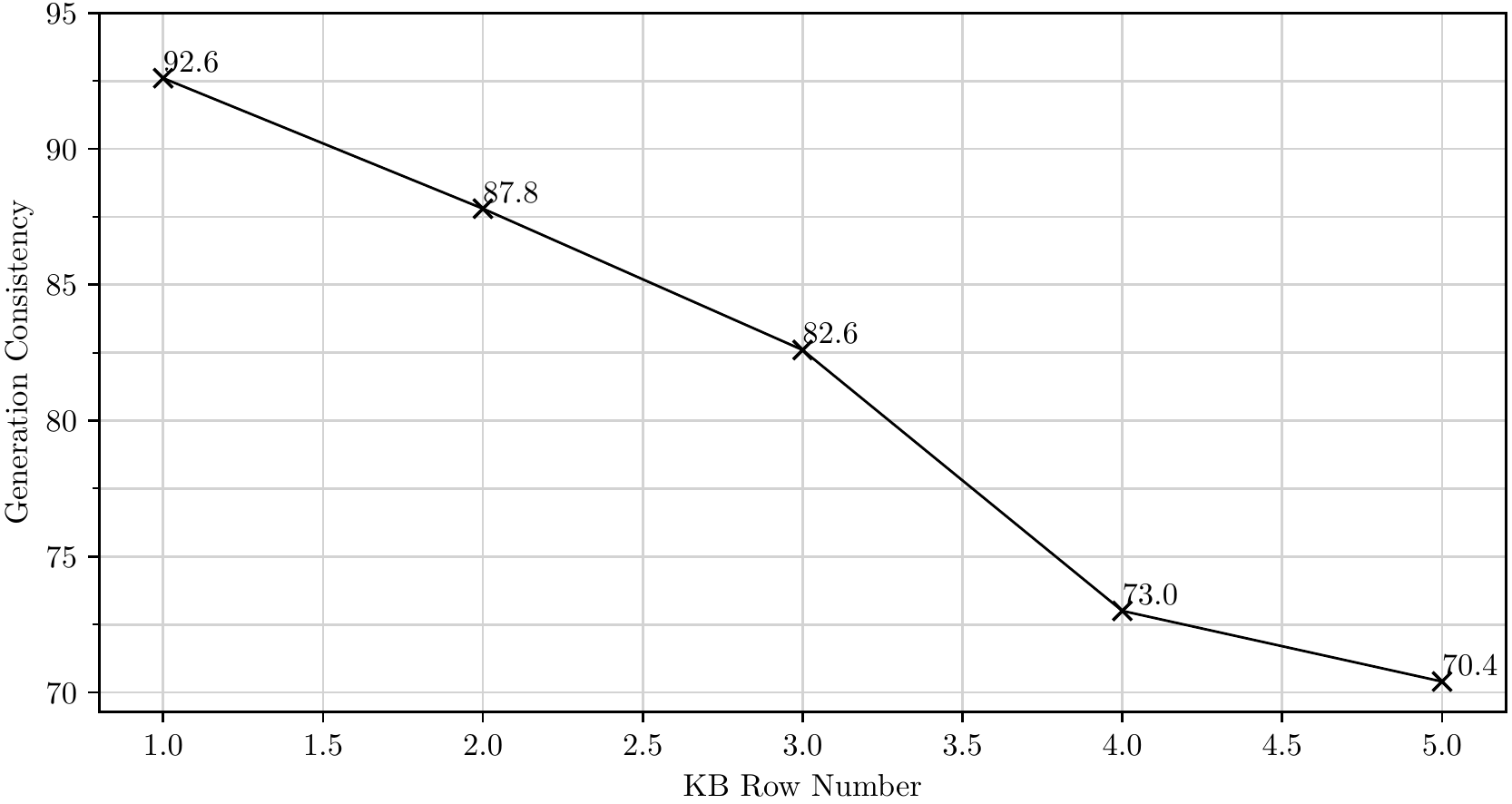}
	\caption{Correlation between the number of KB rows and generation consistency on navigation domain.}\label{fig:row_num}
\end{figure}

\subsection{{Generation Consistency}}
In this paper,
we expect the consistent generation
from our model.
To verify this, we compute the consistency recall
of the utterances that have multiple entities.
An utterance is considered as consistent
if it has multiple entities and these entities belong to the same row which we annotated with distant supervision.

The consistency result is shown in Table \ref{tbl:consistency}.
From this table, we can see that incorporating retriever
in the dialogue generation improves the consistency.

\subsection{{Correlation between the number of KB rows and generation consistency}}
To further explore the correlation between the number of KB rows and generation consistency,
we conduct experiments with distant manner to  study the correlation between the number of KB rows and generation consistency.

We choose KBs with different number of rows on a scale from 1 to 5  for the generation.
From Figure~\ref{fig:row_num}, as the number of KB rows increase, 
we can see a decrease in generation consistency.
This indicates that irrelevant information would harm the dialogue generation consistency.

 	\subsection{Visualization}
\begin{figure}[t]
	\centering

	\includegraphics[scale=0.4]{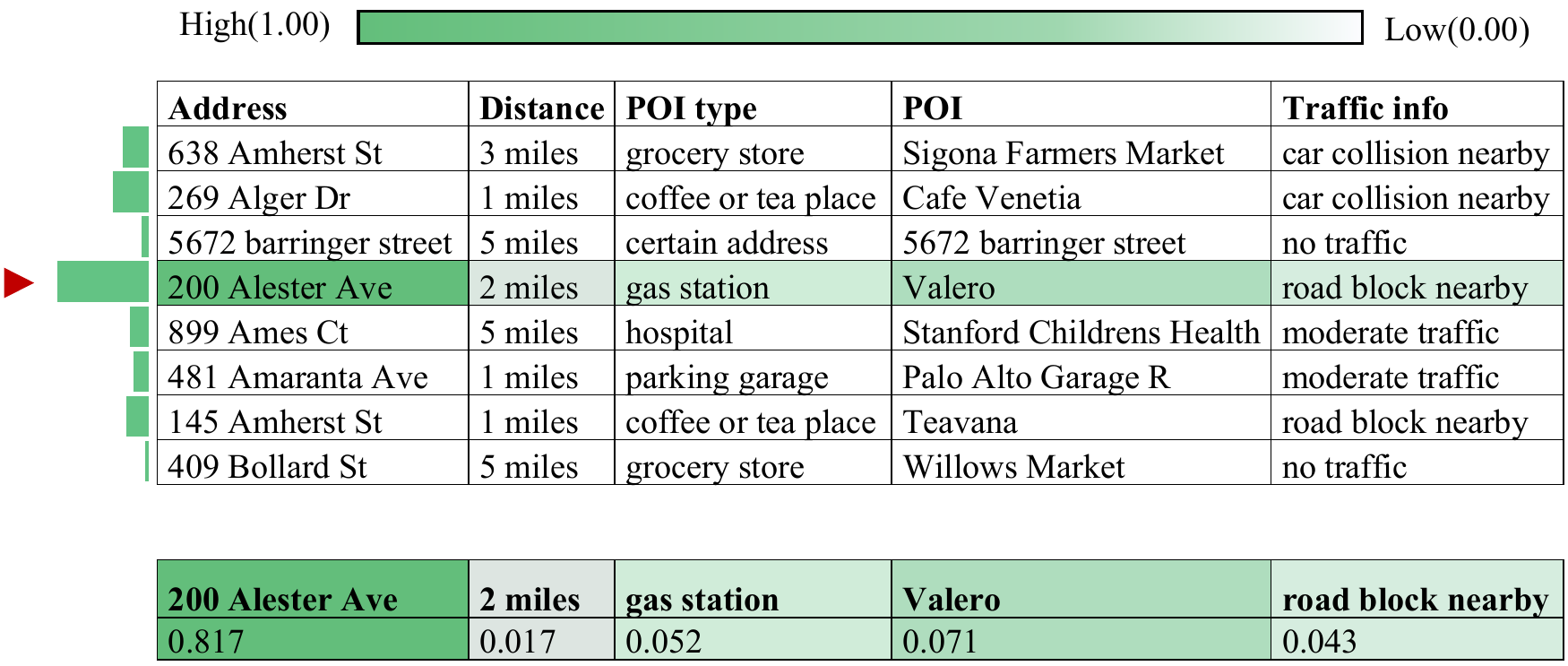}
	\caption{
		KB score distribution.
		The distribution is the timestep when generate entity \texttt{200\_Alester\_Ave} for response ``	\textit{Valero is located at 200\_Alester\_Ave}''
	}
	\label{fig:visual}
\end{figure}

To gain more insights into how the our retriever module influences the whole KB score distribution, we visualized the KB entity probability at the decoding position where we generate the entity \texttt{200\_Alester\_Ave}.
From the example (Fig~\ref{fig:visual}), we can see the $4^\text{th}$ row and the $1^\text{th}$ column has the highest probabilities for generating \texttt{200\_Alester\_Ave}, 
which verify the effectiveness of  firstly selecting the most relevant KB row and further selecting
the most relevant KB column.

 \subsection{Human Evaluation}
 		We provide human evaluation on our framework 
 	and the compared models.
 	These responses are based on distinct dialogue history.
 	We hire several human experts and ask them to judge
 	the quality of the responses according to correctness, fluency, and humanlikeness on a scale from 1 to 5. 
 	In each judgment, the expert is presented
 	with the dialogue history,
 	an output of a system with the name anonymized,
 	and the gold response.

 	The evaluation results are illustrated in Table \ref{tbl:consistency}. 
 	Our framework 
 	outperforms other baseline models on all metrics according to Table \ref{tbl:consistency}.
 	The most significant improvement is from correctness, 
 	indicating that our model can retrieve accurate entity from KB and generate more informative information that the users want to know.
 	\begin{table}[t]
 		
 		\centering
 			\begin{adjustbox}{width=0.4\textwidth}
 		\begin{tabular}{r|c||ccc}
 			\hline
 			
 			\multirow{2}{*}{Model}&
 			\multirow{2}{*}{Cons.}&
 			\multicolumn{3}{c}{Human Evaluation} \\
 			
 			\cline{3-5}
 			& & Cor. & Flu. & Hum. \\
 			
 			\hline
 			Copy Net & 21.2 &4.14&4.40&4.36\\
 			Mem2Seq & 38.1& 4.29& 4.29& 4.27 \\
 			DSR & 70.3& 4.59& 4.71& 4.65 \\
 			\hline
 			w/ distant supervision& 65.8& 4.53 & 4.71 & 4.64  \\
 			w/ Gumble-Softmax & \textbf{72.1}&\textbf{4.64}& \textbf{4.73}& \textbf{4.69} \\
 			\hline
 		\end{tabular}
 	\end{adjustbox}
 		\caption{The generation consistency and Human Evaluation on navigation domain. \textit{Cons.} represents \textit{Consistency}. \textit{Cor.} represents \textit{Correctness}. \textit{Flu}. represents \textit{Fluency} and \textit{Hum.} represents 
 			\textit{Humanlikeness.}}\label{tbl:consistency}
 	\end{table}

\section{Related Work}
Sequence-to-sequence (Seq2Seq) models 
in text generation \cite{sutskever2014sequence,bahdanau2014neural,luong-pham-manning:2015:EMNLP,K16-1028,nallapati2016sequence} has gained more popular and they are applied for the open-domain dialogs \cite{vinyals2015neural, serban2016building} in the end-to-end training method. Recently, the Seq2Seq can be used for learning task oriented dialogs and
how to query the structured KB is the remaining challenges.

Properly querying the KB has long been
a challenge in the task-oriented dialogue system.
In the pipeline system, the KB query
is strongly correlated with the design
of language understanding, state tracking, and policy management.
Typically, after obtaining the dialogue state,
the policy management module issues an API call accordingly
to query the KB.
With the development of neural network in natural language processing,
efforts have been made to replacing the
discrete and pre-defined dialogue state
with the distributed representation
\cite{bordes-weston:2017:ICLR,wen:2017:EACL,wen:2017:ICML,liu:2017:interspeech}.
In our framework, our retrieval result can be treated
as a numeric representation of the API call return. 

Instead of interacting with the KB via API calls,
more and more recent works tried to incorporate KB query
as a part of the model.
The most popular way of modeling KB query
is treating it as an attention network over the entire KB entities \cite{eric:2017:SIGDial,dhingra:2017:ACL,reddy2018multi,raghu-etal-2019-disentangling,wu2019global}
and the return can be a fuzzy summation of the entity representations.
\newcite{madotto2018mem2seq}'s practice of
modeling the KB query with memory network can also be considered
as learning an attentive preference over these entities.
\newcite{wen2018sequence} propose the implicit dialogue state representation to query the KB and achieve the promising performance.
Different from their modes, we propose the KB-retriever to explicitly query the KB, and the query result is used to filter the irrelevant entities in the dialogue generation to improve the consistency among the output entities.
\section{Conclusion}

In this paper, we propose a novel framework to improve entities consistency by querying KB in two steps.
In the first step, inspired by the observation that a response can usually be supported by a single KB row, we introduce the KB retriever to return the most relevant KB row, which is used to filter the irrelevant KB entities and encourage consistent generation.
In the second step, we further perform attention mechanism to select the most relevant KB column.
Experimental results show the effectiveness of our method.
Extensive analysis further confirms the observation
and reveal the correlation between the success of KB query
and the success of task-oriented dialogue generation.

\section*{Acknowledgments}
We thank the anonymous reviewers for their helpful comments and suggestions.
This work was supported by the National Natural Science Foundation of China (NSFC) via grant 61976072, 61632011 and 61772153.

\bibliography{emnlp-ijcnlp-2019}
\bibliographystyle{acl_natbib}

\appendix

\end{document}